\documentclass{article}

\usepackage[bottom]{footmisc}
\usepackage{float}
\usepackage{sidecap}
\usepackage{multirow}
\usepackage{parskip}

\usepackage[margin=1.25in]{geometry}
\usepackage[utf8]{inputenc} 
\usepackage[T1]{fontenc}    
\usepackage{url}            
\usepackage{booktabs}       
\usepackage{amsfonts}       
\usepackage{nicefrac}       
\usepackage{microtype}      
\usepackage{kpfonts}
\usepackage{graphicx}
\usepackage{multicol}
\usepackage{enumitem}

\usepackage[small]{caption}
\usepackage{subcaption}

\usepackage[table]{xcolor}
\definecolor{codegreen}{rgb}{0,0.6,0}
\definecolor{codegray}{rgb}{0.3607843137,
0.4823529412,
0.5725490196}
\definecolor{codeblue}{rgb}{0,0.28,0.67}
\definecolor{backcolour}{rgb}{0.9882352941,
0.9725490196,
0.9294117647}

\usepackage{hyperref}
\hypersetup{
    colorlinks,
    linkcolor={codeblue},
    citecolor={codeblue},
    urlcolor={codeblue}
}

\usepackage[scaled]{beramono}

\usepackage{lipsum}
\usepackage{listings}

\lstdefinestyle{mystyle}{
    escapechar=\%,
    backgroundcolor=\color{backcolour},   
    commentstyle=\color{codegray},
    keywordstyle=\color{codeblue},
    basicstyle=\ttfamily\small,
    breakatwhitespace=false,         
    breaklines=true,                 
    captionpos=b,                    
    keepspaces=true,                 
    numbers=left,                    
    numbersep=5pt,                  
    showspaces=false,                
    showstringspaces=false,
    showtabs=false,                  
    tabsize=2
}
\usepackage{amsthm}

\usepackage{setspace}

\title{\Large
\vspace{-2ex}
End-to-End Test-Time Training for Long Context
}
\author{
\small
Arnuv Tandon\thanks{
\,Core contributors. See statement of contributions before references.\\
Correspondence to:
\url{arnuv@stanford.edu}, 
\url{kdalal@berkeley.edu},
\url{yusun@cs.stanford.edu}. \\
}$~\,^{1,3}$,
Karan Dalal$^*$$^{1,4}$,
Xinhao Li$^*$$^5$,
Daniel Koceja$^*$$^3$, 
Marcel R{\o}d$^*$$^3$, 
Sam Buchanan$^4$,\\
\small
Xiaolong Wang$^5$,
Jure Leskovec$^3$, 
Sanmi Koyejo$^3$, 
Tatsunori Hashimoto$^3$, 
Carlos Guestrin$^3$, \\
\small
Jed McCaleb$^1$,
Yejin Choi$^{2}$,
Yu Sun$^*$$^{2,3}$\\
\small
$^1$\,Astera Institute\quad 
~$^2$\,NVIDIA\quad
~$^3$\,Stanford University\quad
~$^4$\,UC Berkeley\quad
~$^5$\,UC San Diego
\vspace{-0.5ex}
}
\date{}

\begin{document}

\maketitle

\begin{abstract}
\noindent
We formulate long-context language modeling as a problem in continual learning rather than architecture design.
Under this formulation, we only use a standard architecture -- a Transformer with sliding-window attention.
However, our model continues learning at test time via next-token prediction on the given context, compressing the context it reads into its weights.
In addition, we improve the model's initialization for learning at test time via meta-learning at training time.
Overall, our method, a form of Test-Time Training (TTT), is End-to-End (E2E) both at test time (via next-token prediction) and training time (via meta-learning), in contrast to previous forms.
We conduct extensive experiments with a focus on scaling properties.
In particular, for 3B models trained with 164B tokens, our method (TTT-E2E) scales with context length in the same way as Transformer with full attention, while others, such as Mamba\,2 and Gated DeltaNet, do not.
However, similar to RNNs, TTT-E2E has constant inference latency regardless of context length, making it $2.7\times$ faster than full attention for 128K context.
Our \href{https://github.com/test-time-training/e2e}{code} is publicly available.
\end{abstract}

\begin{figure}[h!]
    \centering
    \includegraphics[width=0.495\linewidth]{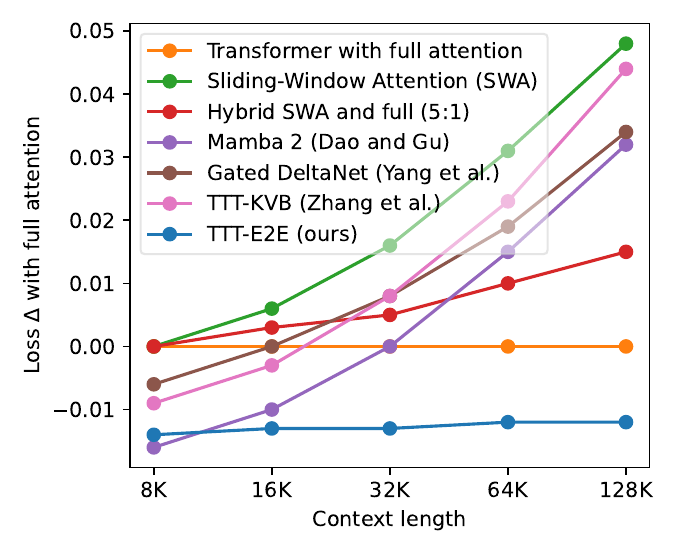}
    \includegraphics[width=0.495\linewidth]{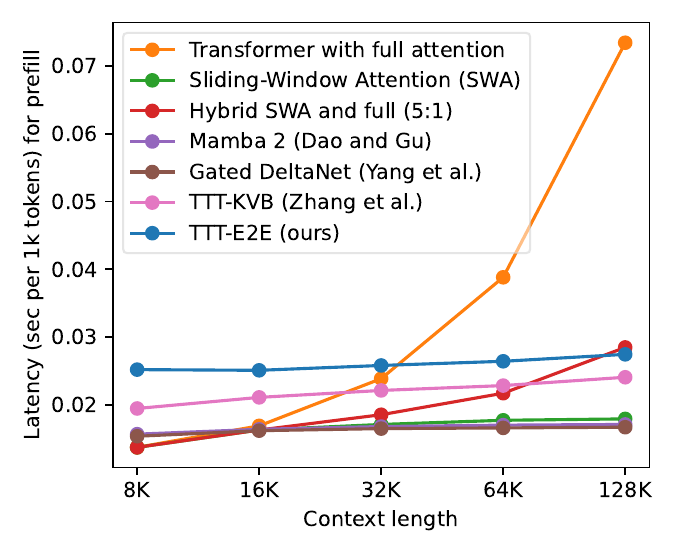}
    \caption{Scaling with context length, in terms of test loss (left) and latency (right).
    \textbf{Left:} 
    Our method (TTT-E2E) turns the worst line (green) into the best (blue) at 128K context length.
    Loss $\Delta$ ($\downarrow$), the $y$-value, is computed as
    (loss of the reported method) $-$ (loss of Transformer with full attention), so loss $\Delta$ of full attention itself (orange) is the flat line at $y=0$.
    While other methods produce worse loss $\Delta$ in longer context, TTT-E2E maintains the same advantage over full attention.
    All models have 3B parameters and are trained with 164B tokens.
    \textbf{Right:}
    Similar to SWA and the RNN baselines, TTT-E2E has constant inference latency regardless of context length, making it $2.7\times$ faster than full attention for 128K context on an H100.
    }
    \label{fig:teaser}
\end{figure}

\clearpage
\section{Introduction}
\label{sec:intro}

Humans are able to improve themselves with more experience throughout their lives, despite their imperfect recall of the exact details.
Consider your first lecture in machine learning: You might not recall the instructor's first word during the lecture, but the intuition you learned is probably helping you understand this paper, even if that lecture happened years ago.

On the other hand, Transformers with self-attention still struggle to efficiently process long context equivalent to years of human experience, in part because they are designed for nearly lossless recall.
Self-attention over the full context, also known as full attention, must scan through the keys and values of all previous tokens for every new token.
As a consequence, it readily attends to every detail, but its cost per token grows linearly with context length and quickly becomes prohibitive.

As an alternative to Transformers,
RNNs such as Mamba\,2~\cite{gu2023mamba} and Gated DeltaNet~\cite{yang2024gated} have constant cost per token, but become less effective in longer context, as shown in Figure~\ref{fig:teaser}.
Some modern architectures approximate full attention with a sliding window~\cite{agarwal2025gpt,yuan2025native}, or stack attention and RNN layers together~\cite{team2025kimi, blakeman2025nemotron}.
However, these techniques are still less effective than full attention in using longer context to achieve better performance in language modeling.

How can we design an effective method for language modeling with only constant cost per token?
Specifically, how can we achieve better performance in longer context without recalling every detail, as in the opening example?
The key mechanism is compression.
For example, humans compress a massive amount of experience into their brains, which preserve the important information while leaving out many details.
For language models, training with next-token prediction also compresses a massive amount of data into their weights.
So what if we just continue training the language model at test time via next-token prediction on the given context?

This form of Test-Time Training (TTT), 
similar to an old idea known as dynamic evaluation~\cite{mikolov2013efficient, krause2018dynamic}, still has a missing piece:
At training time, we were optimizing the model for its loss out of the box, not for its loss after TTT.
To resolve this mismatch, we prepare the model's initialization for TTT via meta-learning~\cite{hinton1987using, schmidhuber1992learning, kirsch2021meta} instead of standard pre-training.
Specifically, each training sequence is first treated as if it were a test sequence, so we perform TTT on it in the inner loop.
Then we average the loss after TTT over many independent training sequences, and optimize this average w.r.t.\ the model's initialization for TTT through gradients of gradients in the outer loop~\cite{maclaurin2015gradient, andrychowicz2016learning, finn2017model}.

In summary, our method is end-to-end in two ways.
Our inner loop directly optimizes the next-token prediction loss at the end of the network, in contrast to prior work on long-context TTT~\cite{sun2023learning, zhang2025test}; 
Subsection~\ref{subsec:alternative} explains this difference through an alternative derivation of our method.
Moreover, our outer loop directly optimizes the final loss after TTT, in contrast to dynamic evaluation~\cite{mikolov2013efficient, krause2018dynamic}, as discussed.
Our key results are highlighted in Figure~\ref{fig:teaser}, with the rest presented in Section~\ref{sec:results}.

The conceptual framework of TTT has a long history with many applications beyond long context, and many forms without meta-learning~\cite{stone1977consistent, bottou1992local, hubotter2024efficiently, akyurek2024surprising}.
Our work is also inspired by the literature on fast weights~\cite{hinton1987using, schmidhuber1992learning, schlag2020learning, irie2021going}, especially \cite{clark2022meta} by Clark et al., which shares our high-level approach.
Section~\ref{sec:related} discusses related work in detail.

\section{Method}
\begin{figure}
\vspace{-6ex}
    \centering
    \includegraphics[width=0.485\linewidth]{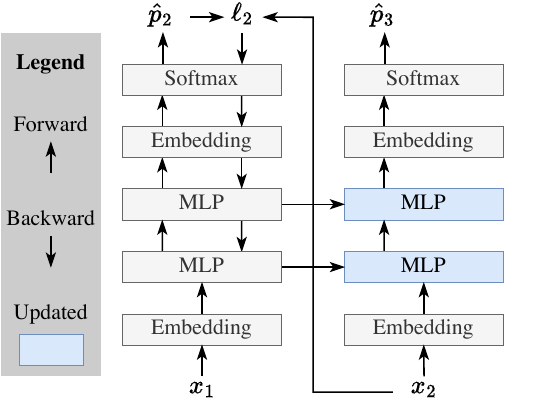}
    \includegraphics[width=0.495\linewidth]{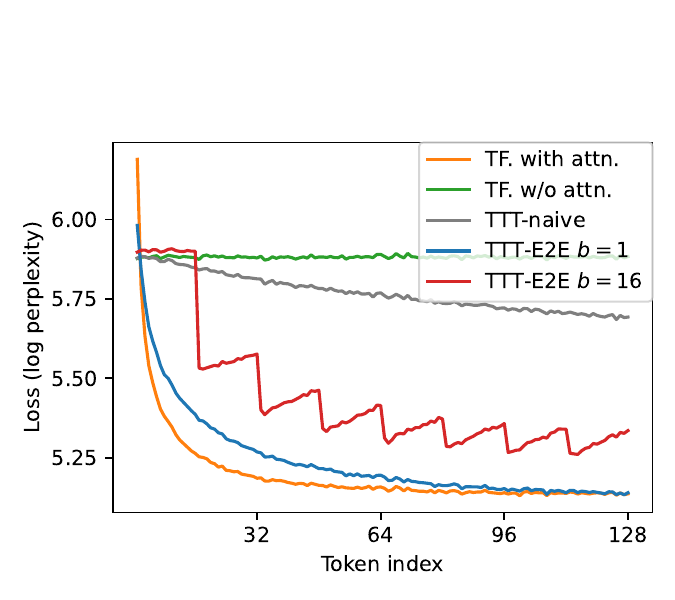}
    \caption{Toy example.
    \textbf{Left:} Given $x_1$ and $x_2$ as context, we want to predict the unknown $x_3$.
    Our toy baseline, a Transformer without self-attention (using only the upward arrows), is effectively a bigram since it has no memory of $x_1$.
    TTT (using all the arrows) first tries to predict $x_2$ from $x_1$ as an exercise: It computes the loss $\ell_2$ between $x_2$ and the prediction $\hat{p}_2$, then takes a gradient step on $\ell_2$.
    Now information of $x_1$ is stored in the updated MLPs (blue).
    \textbf{Right:} Token-level test loss $\ell_t$ for various methods in our toy example, as discussed in Subsection~\ref{subsec:meta}, except for TTT-E2E $b=16$ discussed in Subsection~\ref{subsec:main}.
    In particular, TTT-E2E $b=1$ turns the green line (our toy baseline) into the blue line, which performs almost as well as orange (using full attention).
    }
    \label{fig:toy}
\end{figure}

Consider the standard task of next-token prediction, which consists of two phases at test time:
\vspace{-0.5ex}
\begin{itemize}[itemsep=2pt, topsep=0pt, parsep=0pt, partopsep=0pt]
    \item Prefill: conditioning on $T+1$ given tokens $x_0, x_1, \dots, x_T$, where $x_0$ is the  Beginning of Sequence (\texttt{<BOS>}) token.
    \item Decode: predicting a distribution $\hat{p}_{T+1}$ over all possible instantiations of the next token.
\end{itemize}
\vspace{-0.5ex}
The test loss is then $\texttt{CE}\left(\hat{p}_{T+1}, x_{T+1}\right)$, where \texttt{CE} is the cross entropy and $x_{T+1}$ is generated by nature.

For ease of exposition, we first focus on the task of prefilling $T+1$ tokens and then decoding a single token.
In this setting, self-attention over the full context, also known as full attention, has computational complexity $O(T^2)$ for prefill and $O(T)$ for decode.
We now discuss our method using Test-Time Training (TTT), which has $O(T)$ for prefill and $O(1)$ for decode.

\subsection{TTT via Next-Token Prediction}
\label{subsec:toy}

To motivate our main method, we introduce a toy example that we will develop all the way up to the middle of Subsection~\ref{subsec:main}.
This toy example is based on a rather silly architecture: a Transformer with all of its self-attention layers removed, leaving only the MLP layers.
Our toy baseline -- blithely applying this architecture to language modeling -- is effectively a bigram since it has no memory of previous tokens.
Our goal is to understand the effect of TTT in isolation, without the confounding of other sequence modeling components.

One way to give our baseline architecture some memory is to train it on the context.
Similar to standard pre-training, we can predict $\hat{p}_t$ and compare it to $x_t$ at every $t = 1, \dots, T$ as an exercise.
Specifically, denote the baseline architecture as $f$ with weights $W$, then the standard next-token prediction loss at time $t$ can be written as:
\begin{equation}
\ell_t(W) = \texttt{CE}\left(f(x_{t-1}; W), x_t\right).
\end{equation}
We update $W$ at test time for every $t=1,\dots,T$, in sequential order, with gradient descent:
\begin{equation}
\label{eq:toy}
W_t = W_{t-1} - \eta\,\nabla\ell_t(W_{t-1}),
\end{equation}
where $\eta$ is a learning rate, and $W_0$ is the initial weights at test time.
In the end, we simply output 
$\hat{p}_{T+1} = f(x_T; W_T)$.
We illustrate this form of TTT in the left panel of Figure~\ref{fig:toy}:
Our toy baseline only uses the upward arrows, while TTT adds the backward and horizontal arrows.

\subsection{Learning to (Learn at Test Time)}
\label{subsec:meta}

We now consider how $W_0$ -- the initial weights after (training-time) training but before test-time training -- is obtained, within the scope of the same toy example.
By definition, our test-time training loss $\ell_t(W_{t-1})$ 
is also the test loss for the next-token prediction task that conditions on $x_0,\dots,x_{t-1}$ and tries to predict $x_t$. 
Therefore, the test loss over a sequence $X = (x_1, \dots, x_T)$ is:
\begin{equation}
\label{eq:toy_train}
\mathcal{L}(W_0; X) = \frac{1}{T}\sum_{t=1}^{T}\ell_t(W_{t-1}) = \frac{1}{T}\sum_{t=1}^{T} \texttt{CE}\left(f(x_{t-1}; W_{t-1}), x_t\right).
\end{equation}
To obtain a $W_0$ that will produce low $\mathcal{L}(W_0; X)$ at test time, the most direct approach is to optimize the same loss at training time over a large training set of sequences on average.
This direct approach is an example of End-to-End (E2E) training, where the training loss matches the test loss.
When TTT uses a $W_0$ trained in this fashion, we call it \emph{TTT-E2E}.

As a contrasting example, consider another approach that naively imitates the training loss of a static model without taking into account that $W_0$ will be updated at test time:
\begin{equation}
\label{eq:wrong}
\mathcal{L}_{\texttt{naive}}(W_0; X) = \frac{1}{T}\sum_{t=1}^{T}\ell_t(W_0).
\end{equation}
This approach is not E2E, since there is a mismatch between the model's behavior at training and test time. As a consequence, we can provide little guarantee that a minimizer of $\mathcal{L}_{\texttt{naive}}$ will also produce low test loss $\mathcal{L}$.
We call this approach \emph{TTT-naive}.
It has been the mainstream approach in the literature of dynamic evaluation~\cite{mikolov2013efficient, krause2018dynamic}.

The right panel of Figure~\ref{fig:toy} plots the token-level test loss $\ell_t$, averaged over many test sequences, for $t=1,\dots,128$. So far, we have discussed four methods: Transformer with full attention (orange), our toy baseline without attention (green), TTT-naive (gray), and TTT-E2E (blue for $b=1$; we will cover the variant with $b=16$ in Subsection~\ref{subsec:main}); see details of the experimental setup in Appendix~\ref{app:toy}.
While TTT-naive performs only slightly better than the toy baseline, 
TTT-E2E performs almost as well as full attention.
In particular, TTT-E2E can effectively use more context to better predict the next token,
as demonstrated by the test loss decreasing over time.

For gradient-based optimization, computing $\nabla \mathcal{L}(W_0)$ for the E2E $\mathcal{L}$ entails computing gradients of gradients, since the update rule in Equation~\ref{eq:toy} itself contains a gradient operation.
Fortunately, modern frameworks for automatic differentiation can efficiently compute gradients of gradients with minimal overhead~\cite{jax2018github, engstrom2025optimizing}.
Once $\nabla \mathcal{L}(W_0)$ is computed, we can plug it into standard optimizers.
In the field of meta-learning, gradient steps on $\mathcal{L}$ are called the \emph{outer loop}, and on $\ell$ the \emph{inner loop}.

The current version of TTT-E2E still has two problems for large models in long context.
The first is efficiency, because our inner loop has many steps that cannot be parallelized.
The second is stability, because each gradient step in the inner loop depends on only a single token, which can easily lead to gradient explosion by chance.
The next subsection addresses these two problems.

\subsection{Mini-Batch TTT and Sliding Window}
\label{subsec:main}

The two problems above share a common cause: Equation~\ref{eq:toy} performs online instead of mini-batch gradient descent.
Given a training set of size $T$, the standard practice is to partition it into $T/b$ batches, each of size $b$ (assuming divisible), and take one gradient step per batch.
Compared to online gradient descent, where $b=1$, a larger $b$ is known to improve both parallelism and stability.
We can apply the mini-batch idea to TTT for the same benefits.
Given the (test-time) training set that contains $x_1,\dots,x_T$,  
we generalize Equation~\ref{eq:toy} to:
\vspace{-0.5ex}
\begin{equation}
\label{eq:main}
W_i = W_{i-1} - \eta\,\frac{1}{b}\sum_{t=(i-1)b+1}^{ib}\nabla\ell_t(W_{i-1}),
\end{equation}
for $i=1,\dots,T/b$ (assuming divisible), then output $\hat{p}_{T+1} = f(x_T; W_{T/b})$.
In addition, for (training-time) training to reflect the change in test-time training, we also generalize Equation~\ref{eq:toy_train} to:
\begin{equation}
\label{eq:main_train}
\mathcal{L}(W_0; X) = \frac{1}{T}\sum_{i=1}^{T/b}\,\sum_{t=(i-1)b+1}^{ib}\ell_t(W_{i-1}).
\vspace{-0.5ex}
\end{equation}
Note that $b=1$ recovers Equation~\ref{eq:toy} and~\ref{eq:toy_train}.

However, our model with mini-batch TTT is now a bigram again within each batch, as illustrated by the 
red line in Figure~\ref{fig:toy} with $b=16$. 
For example, consider the first mini-batch that contains $x_1,\dots,x_b$.
Since every prediction $\hat{p}_t = f(x_{t-1}; W_0)$ is made with $W_0$ instead of $W_{t-1}$, we observe that $\ell_t(W_0)$ increases with $t$ as $\hat{p}_t$ misses more context, namely all the tokens up to $t-1$.
This observation holds within every mini-batch, where the only predictions without missing context are the first and second ones inside the mini-batch.
These increasing losses produce worse gradient steps for TTT, which ultimately translate into worse performance of the purple line compared to the blue line.

To address this problem, we finally advance beyond the toy example and augment our architecture with sliding-window attention layers.
While our toy example removes the self-attention layers entirely, 
our main method only restricts them to a fixed window size $k$.
For our main results with $T=128$K, we set the window size $k$ to 8K and the TTT mini-batch size $b$ to 1K.
It is important to set $k\geq b$ so our model can remember the context within each mini-batch before TTT has a chance to update its weights.

\begin{figure}[t!]
    \centering
    \vspace{-3ex}                                                          
    \includegraphics[width=0.95\linewidth]{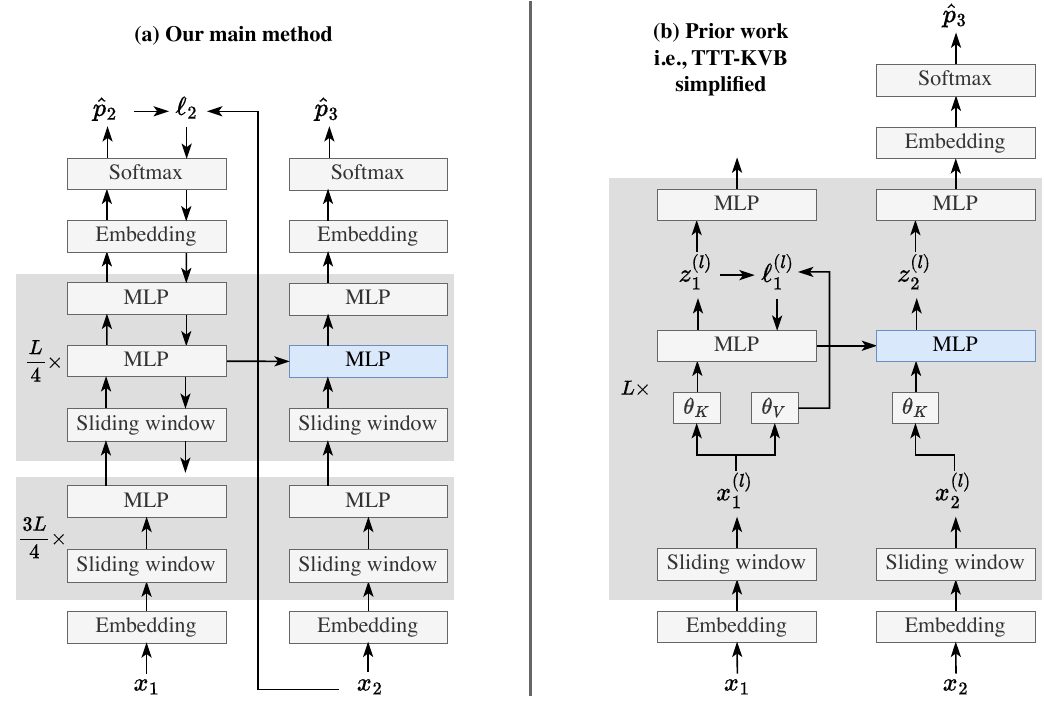}
    \caption{
    Computation graphs following the setup in Figure~\ref{fig:toy}: 
    Given $x_1$ and $x_2$ as context, we want to predict the unknown $x_3$. 
    \textbf{Left:} 
    Our main method with the sliding-window attention layers and the implementation details discussed in Subsection~\ref{subsec:main}.
    For ease of notation, our illustration uses online gradient descent ($b=1$).
    The lowest downward arrow is disconnected to the MLP below, since gradients pass through the last $L/4$ blocks but not further down. 
    \textbf{Right:}
    The first step of our alternative derivation in Subsection~\ref{subsec:alternative}:
    a simplified version of TTT-KVB in prior work~\cite{zhang2025test, sun2024learning}.
    }
    \label{fig:main}
\end{figure}

This modification of the baseline architecture completes our main method.
Next, we introduce three implementation details, and then consider the task of decoding multiple tokens. Our main method, complete with the implementation details, is illustrated in the left panel of Figure~\ref{fig:main}.

\vspace{-0.5ex}
\subsubsection{Implementation Details}
Three implementation details are necessary for achieving our reported results.
We will justify these details with ablations in Section~\ref{sec:results}.
However, it is still possible that they are merely artifacts of our experimental setup, and different design choices could be better suited in other setups.

\textbf{TTT only the MLP layers.}
Modern Transformers are built in repeated blocks, each consisting of a full attention layer (which we have replaced with sliding window attention), an MLP layer, and a few normalization layers.
We freeze the embedding layers, normalization layers, and attention layers during TTT, since updating them in the inner loop causes instability in the outer loop.
Therefore, the MLP layers are the only ones updated during TTT. 

\textbf{TTT only 1/4 of the blocks.}
In general, less information is lost during compression when we have a larger amount of storage.
In our case, the information is the context, and the storage is the updated MLP layers.
However, updating more layers also implies more computation to back-propagate the gradients.
Therefore, we have an intuitive trade-off between computational cost and the ability to scale with context length, as we will illustrate with ablations in Section~\ref{sec:results}.
We choose to TTT only the last 1/4 of the blocks according to the ablations, but other experimental setups, especially those with even longer contexts, might require a different choice.

\textbf{Two MLP layers per block.}
One of the concerns of TTT is forgetting the knowledge learned during pre-training.
We adopt the simplest way to address this concern.
In the blocks updated during TTT, we add a static, second MLP layer 
as a ``safe'' storage for pre-trained knowledge.
For fair comparison with the baselines, we reduce the hidden dimension of the MLPs throughout the entire network (including those frozen during TTT), so the total number of parameters remains the same.

\subsubsection{Decoding Multiple Tokens}

Up to this point, we have focused on the task of prefilling $T+1$ tokens (including \texttt{<BOS>} as $x_0$) and then decoding a single token.
We now consider decoding multiple tokens, for which our method admits a natural extension:
It only takes a gradient step once the decoded tokens have completely filled a TTT mini-batch.
For example, assuming that $T$ is divisible by $b$, so TTT depletes the prefilled tokens in exactly $T/b$ mini-batches.
Then our method does not need to do anything special when decoding the next $b$ tokens. 
After that, it performs TTT on this batch of decoded tokens, and then continues to decode using the updated weights.

\renewcommand{\arraystretch}{1.5}
\begin{table}[t]
\vspace{-2ex}
\centering
\begin{minipage}{0.45\textwidth}
\scalebox{0.92}{
\begin{tabular}{|l|c|c|}
\hline
\textbf{Method} & \textbf{Loss} & \textbf{Diff.} \\
\hline
SWA ($k=8$K) baseline   & 2.827 & - \\
\hline
TTT-KVB (Zhang et al.)  & 2.818 & $-$0.009 \\
TTT-KVB simplified      & 2.819 & +0.001 \\
TTT-E2E all layers MH   & 2.806 & $-$0.013 \\
TTT-E2E (ours)          & 2.805 & $-$0.001 \\
\hline
\end{tabular}
}
\end{minipage}
\hspace{2ex}
\begin{minipage}{0.5\textwidth}
\vspace{2ex}
\caption{
Our alternative derivation in Subsection~\ref{subsec:alternative} starts from TTT-KVB (Zhang et al.~\cite{zhang2025test}) and takes three steps to reach TTT-E2E (ours).
Here, we present results along these steps, as well as the result of SWA as a point of reference.
We consider a difference of 0.001 below the threshold of statistical significance.
For the results here,
we pre-trained and evaluated 760M models on DCLM with our default hyper-parameters, following the basic recipe in Section~\ref{sec:results}.
MH stands for multi-head.
}
\label{tab:ablations}
\end{minipage}
\vspace{-2ex}
\end{table}

\subsection{Alternative Derivation}
\label{subsec:alternative}

This subsection discusses an alternative derivation of our main method, starting from prior work on long-context TTT based on Key-Value Binding (KVB)~\cite{sun2024learning, zhang2025test}.
The key step is to replace their layer-wise reconstruction loss with the standard next-token prediction loss, so TTT becomes E2E at test time. 
This derivation is not needed to understand the results in Section~\ref{sec:results}, but it provides additional insight into how our method is connected to the literature on RNNs.

\subsubsection{Starting Point: Key-Value Binding}

Building on the same idea of compressing context into the weights of a model, prior work~\cite{sun2024learning} uses TTT to construct a sequence modeling layer that serves as a drop-in replacement for self-attention.
While self-attention associates the key and value of every previous token by storing them explicitly in a cache, prior work proposes storing these associations implicitly in a model, by learning at test time to predict each value from its key.
This learning objective, later known as KV Binding, has been the core component in many popular variants of TTT, such as MesaNet~\cite{von2025mesanet}, Titans~\cite{behrouz2024titans}, and Nested Learning~\cite{behrouz2025nested}; 
linear attention~\cite{schmidhuber1992learning, katharopoulos2020transformers} and many of its variants, such as Gated DeltaNet~\cite{yang2024gated}, can also be derived from this perspective.\footnote{
\setstretch{1.25}
As shown in~\cite{sun2024learning}, 
when $g$ is a linear model, TTT-KVB recovers DeltaNet~\cite{schlag2021linear};
when the gradient in Equation~\ref{eq:multi} is taken w.r.t. $W^{(l)}_0$ instead of $W^{(l)}_{t-1}$, TTT-KVB recovers linear attention~\cite{schmidhuber1992learning, katharopoulos2020transformers}; 
and when $\ell_t^{(l)}$ is learned using a non-parametric kernel estimator instead of a parametric model, TTT-KVB recovers self-attention.
}

Concretely, given the input embeddings $x_t^{(l)}$ at layer $l$, the basic form of TTT-KVB takes a gradient step at each $t=1,\dots, T$ on the following loss~\cite{sun2024learning}:
\begin{equation}
\label{eq:multi}
\ell_t^{(l)}\left(W_{t-1}^{(l)}\right) = \big\| g\left(\theta_K^{(l)} x_t^{(l)}; W_{t-1}^{(l)}\right) - \theta_V^{(l)} x_t^{(l)} \big\|^2,
\vspace{-1ex}
\end{equation}
where $g$ is usually an MLP, $W_{t-1}^{(l)}$ is the weights of $g$ after the previous timestep, 
and $\theta_K^{(l)}$ and $\theta_V^{(l)}$ are outer-loop parameters, similar to the key-value projection matrices in Transformers.
After the gradient step, $g$ uses the updated weights to produce the output embedding:
\begin{equation}
\label{eq:multi_output}
z_t^{(l)} = g\left(\theta_Q^{(l)} x_t^{(l)}; W_t^{(l)}\right),
\vspace{-1ex}
\end{equation}
where $\theta_Q^{(l)}$ is also a set of outer-loop parameters.
The mechanism above is known as a TTT layer.
When used inside a network, every TTT layer is an independent unit with its own loss and weights.
At training time, the outer loop of TTT-KVB is identical to that of TTT-E2E in Equation \ref{eq:main_train}, 
and all the outer-loop parameters, including $\theta_K$, $\theta_V$, and $\theta_Q$ of all the TTT layers, are optimized together.
Similar to TTT-E2E in Subsection~\ref{subsec:main}, TTT(-KVB) layers can effectively use (inner-loop) mini-batch gradient descent when preceded by sliding-window attention layers~\cite{zhang2025test}.
This hybrid architecture serves as the starting point of our derivation.

\subsubsection{First Step: Simplified Output Rule}

First, we observe that the output rule in Equation~\ref{eq:multi_output} can be simplified into:
\begin{equation}
\label{eq:multi_output_simplified}
z_t^{(l)} = g\left(\theta_K^{(l)} x_t^{(l)}; W_{t-1}^{(l)}\right),
\end{equation}
with practically no harm, as shown in Table~\ref{tab:ablations}.
This new output rule is a simplification because it reuses the prediction of $g$ in Equation~\ref{eq:multi} as the output embedding instead of calling $g$ again with the updated weights and the separate input $\theta_Qx_t$.
Intuitively, calling $g$ with the updated weights can be unnecessary if sliding-window attention already provides enough local context, and prior work has argued that the separation between $\theta_K$ and $\theta_Q$ can also be unnecessary~\cite{kitaev2020reformer, tay2021synthesizer, zhai2021attention}.

In the right panel of Figure~\ref{fig:main}, we illustrate TTT-KVB after this simplification. 
Compared to TTT-E2E in the left panel, there are four differences, 
with the latter two considered implementation details:
\vspace{-0.5ex}
\begin{enumerate}[itemsep=2pt, topsep=0pt, parsep=0pt, partopsep=0pt]
    \item Each Transformer block in TTT-KVB has a reconstruction loss $\ell^{(l)}$, whereas TTT-E2E has a single next-token prediction loss $\ell$ at the end of the entire network.
    \item Each Transformer block in TTT-KVB has additional outer-loop parameters $\theta_K$ and $\theta_V$.
    \item TTT-KVB updates an MLP layer in every Transformer block, whereas TTT-E2E only updates an MLP layer in the last 1/4 of the blocks.
    \item Not shown in the figure, the updated MLPs in TTT-KVB are split into multiple heads in the same way as self-attention, so these MLPs are much smaller than the regular ones in TTT-E2E.\footnote{
    Why is a multi-head MLP smaller than a regular one? First consider a linear model. If an input has $D$ dimensions and we have $H$ heads, then the input to each head will have $D/H$ dimensions. Therefore, the linear model for each head will have $(D/H)^2$ parameters, and all the heads combined will have $D^2/H$ parameters, so more heads will lead to fewer parameters. Since an MLP is simply a stack of linear models, we have a similar relationship.
    }
    Moreover, these MLPs are updated with LoRA~\cite{hu2022lora}, so their effective capacity is even smaller.
\end{enumerate}
\vspace{-0.5ex}
However, Figure~\ref{fig:main} also highlights the similarity between these two forms of TTT:
Similar to TTT-E2E, TTT-KVB can be understood from the perspective of training the entire network.
First, there is a forward pass through the entire network, as illustrated by all the upward arrows.
Then there is a backward pass, with contributions from many losses in the fashion of Deeply Supervised Nets~\cite{lee2015deeply}. However, the gradients are stopped after reaching only one MLP layer, 
as illustrated by the single downward arrow in each block.

\subsubsection{Key Step: E2E at Test Time}

The key step in this derivation is to replace the KVB loss with the next-token prediction loss, which implies removing differences 1 and 2 together, since without the layer-wise reconstruction losses, there is also no use for $\theta_K$ or $\theta_V$.
This step brings us to an intermediate method called \emph{TTT-E2E all layers MH}, where MH stands for multi-head.
As shown in Table~\ref{tab:ablations}, replacing the loss significantly improves performance in language modeling, essentially reaching the level of our final method.

This intermediate method is now E2E at test time, because its (test-time) training loss is exactly the token-level test loss $\ell_t$.
At this point, it is especially interesting to recognize the duality between our two derivations.
Our primary derivation starts from TTT via next-token prediction, which is E2E at test time, and focused on making it E2E at training time via meta-learning in Subsection~\ref{subsec:meta}.
Our alternative derivation, on the other hand, starts from TTT-KVB, which is E2E at training time, and focused on making it E2E at test time via next-token prediction.

\subsubsection{Final Step: Larger State with Less Compute}

TTT(-KVB) layers are often viewed as a class of RNN layers~\cite{sun2024learning},
and TTT-KVB is often viewed as an RNN.\footnote{
Specifically, the hidden state of each TTT layer is the weights of the model $g$, and its update rule and output rule are defined by Equation~\ref{eq:multi} and~\ref{eq:multi_output}.
}
Similarly, TTT-E2E can also be viewed as an RNN, except that it only has one RNN layer, since the entire network is updated in one backward pass.
Among the three components of an RNN, we have modified the output rule in the first step of this derivation, and the update rule in the second (the key) step.
Now we modify the hidden state.

A critical factor that often improves the long-context performance of an RNN is a larger \mbox{hidden state}, which, in turn, often requires more compute to be updated.
Consider our intermediate method, TTT-E2E all layers MH.
If we remove difference 3 by updating only the last 1/4 of the blocks, then we save compute at test time but end up with a smaller state.
And if we remove difference 4 by reverting to regular MLPs (instead of multi-head MLPs with LoRA), then we have a larger effective state at the cost of more compute (and memory).

However, when using the E2E loss, the trade-offs of these two differences are rather disproportionate: In order to update the small multi-head MLP in a block, gradients need to back-propagate through the large MLP above it, let alone the attention layer below for the backward pass to proceed further.
Given the heavy cost of preparing the upstream gradients, it should be more cost-effective to update fewer blocks, each containing a larger hidden state.
Indeed, our final method (TTT-E2E), which removes both differences together, has $5\times$ larger hidden state (88M vs.\,18M for the 760M model) and half the inference latency (0.0086 vs.\,0.017 sec per 1K tokens for prefill on H100) compared to TTT-E2E all layers MH.

Does this larger state actually improve performance? 
It is difficult to see the difference in Table~\ref{tab:ablations} because these experiments are only at 8K context length.
In Subsection~\ref{subsec:results_ablations}, we will investigate the effect of state size in terms of scaling with context length, by ablating the number of layers updated.
This ablation will clearly show that a smaller state leads to worse context scaling.

\section{Main Results}
\label{sec:results}

All experiments can be reproduced using the code and datasets provided in our public repository: {\small\url{https://github.com/test-time-training/e2e}}.
\vspace{-1ex}
\subsection{Setup}
Given the research nature of this paper, our goal is to experiment in the simplest setups at small and medium scales that can inform production-level training runs at large scale.
In general, today's large-scale runs usually consist of two or more stages~\cite{dubey2024llama, xiong2023effective, liu2024deepseek, olmo2025olmo3}:
\vspace{-0.5ex}
\begin{itemize}[itemsep=2pt, topsep=0pt, parsep=0pt, partopsep=0pt]
\item Pre-training at short context length on a general dataset containing diverse knowledge.
\item Extending the context length by fine-tuning on a dataset of long sequences. 
To gradually reach very long context, e.g., 1M, extension is usually broken down further into multiple stages.
\end{itemize}
\vspace{-0.5ex}
For simplicity, our training runs consist of only two stages: pre-training at 8K context length, and extension fine-tuning at the final context length, at most 128K, depending on the experiment.

\textbf{Datasets.}
For pre-training, we use DCLM, specifically DCLM-Baseline, a heavily filtered subset of Common Crawl~\cite{li2024datacomp}.
Given the 3.8T tokens in DCLM-Baseline, we first discard all documents shorter than 8K, our pre-training context length, and then randomly sample from the remaining ones to construct training sets of various sizes.\footnote{
We discard all documents shorter than 8K to avoid resetting the updated MLPs across document boundaries when different documents are packed into the same training sequence, since resetting slows down training for our infrastructure.
}
However, most of the sequences in DCLM that are longer than 128K, our maximum context length for extension, are of low quality. 
So for fine-tuning, we use Books~\cite{gao2020pile}, a standard academic dataset for long-context extension~\cite{authorsguild2023, liu2024world}.
We also use a held-out partition of Books for language modeling evaluation.

\textbf{Basic recipe.}
We experiment with models of five sizes, ranging from 125M to 3B parameters.
Our configurations and training hyper-parameters in various experiments are all derived from a single basic recipe detailed in Appendix~\ref{app:configs}.
In summary, the basic recipe for model configurations and pre-training is taken from GPT-3~\cite{brown2020language} and Mamba~\cite{gu2023mamba};
to produce the basic recipe for fine-tuning, we performed grid search for the Transformer baseline with full attention.

\textbf{Baselines.}
We compare our method with six baselines that represent the state-of-the-art approaches in architecture design.
All the baselines with sliding window use the same window size $k=8$K.
\vspace{-0.5ex}
\begin{enumerate}[itemsep=2pt, topsep=0pt, parsep=0pt, partopsep=0pt]
    \item Transformer with full attention~\cite{touvron2023llama}: with the model configurations discussed above.
    \item Transformer with Sliding-Window Attention (SWA)~\cite{beltagy2020longformer}: with every full attention layer replaced by a SWA layer. Our main method in Subsection~\ref{subsec:main}, without the implementation details, is also based on this architecture.
    The window size $k$ is set to 8K in all our experiments, except for the window size ablations.
    Since the pre-training context length is also 8K, the full attention and SWA baselines are identical until extension fine-tuning.
    \item Hybrid SWA and full attention (5:1)~\cite{team2024gemma}: repeating the pattern of five SWA layers followed by one full attention layer, in the style of Gemma~\cite{team2024gemma}.
    \item Mamba 2~\cite{dao2024transformers}: a popular RNN that uses a hybrid of Mamba 2 layers and SWA layers; tested at large scale in Nemotron-H~\cite{blakeman2025nemotron}.
    \item Gated DeltaNet~\cite{yang2024gated}: a popular RNN that extends Mamba~2 and DeltaNet~\cite{yang2024parallelizing}, and uses a hybrid of Gated DeltaNet layers and SWA layers; tested at large scale in Kimi Linear~\cite{team2025kimi}.
    \item TTT-KVB~\cite{zhang2025test}: a popular RNN that uses a hybrid of TTT-MLP layers with Key-Value Binding (KVB)~\cite{sun2024learning} and SWA layers; also our starting point in Subsection~\ref{subsec:alternative} (without the simplified output rule).
    Titans~\cite{behrouz2024titans} and Nested Learning~\cite{behrouz2025nested} follow a similar construction.
\end{enumerate}
\vspace{-0.5ex}
We implement baselines 1--3 in JAX, together with our own method. 
For baselines 4--6, we use the official code and configurations provided by the authors and have consulted them to improve the baselines when possible. 
Our improvements to the baselines are discussed in Appendix~\ref{app:baselines}.

\begin{figure}
\vspace{-1ex}
    \centering
    \includegraphics[width=\linewidth]{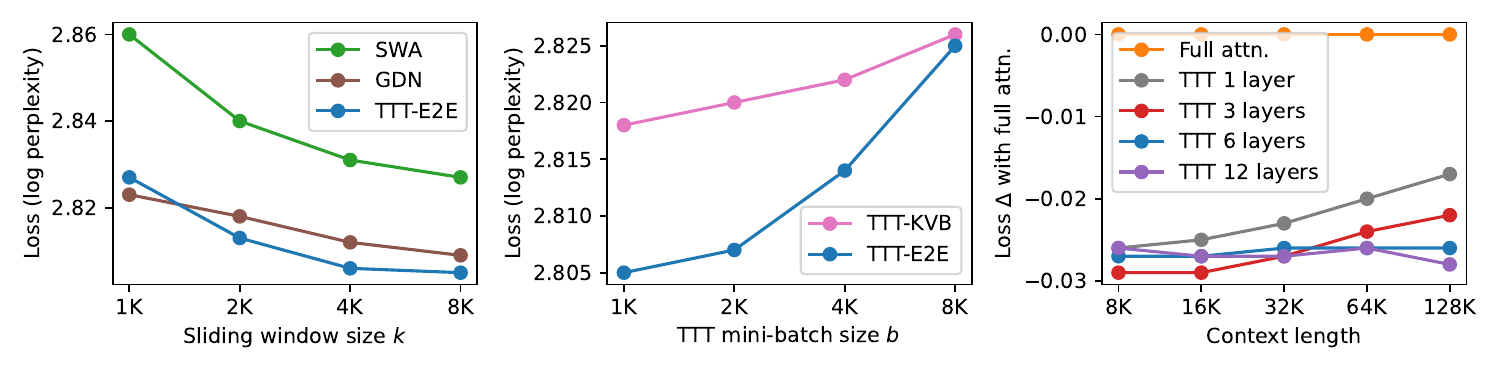}
\vspace{-3ex}
    \caption{
    Ablations on three hyper-parameters: sliding window size $k$, mini-batch size $b$, and the number of layers updated during TTT; see details in Subsection~\ref{subsec:results_ablations}.
    Given the trends in these ablations, we set $k=8$K, $b=1$K, and we update 1/4 the total number of layers.
    Loss $\Delta$ ($\downarrow$), the $y$-value in the rightmost panel, is the same as in Figure~\ref{fig:teaser}. It is computed as (loss of the reported method) $-$ (loss of Transformer with full attention), so loss $\Delta$ of full attention itself (orange) is the flat line at $y=0$.
    GDN stands for Gated DeltaNet~\cite{yang2023gated}.
    }
\vspace{-1ex}
    \label{fig:ablations}
\end{figure}

\subsection{Ablations on Hyper-Parameters}
\label{subsec:results_ablations}

To help readers gradually build an empirical intuition for our method, we start with the simplest experiments -- ablations on the hyper-parameters introduced in Subsection~\ref{subsec:main}.
For all the ablations, we use the 760M model with the basic recipe.

\textbf{Sliding window size $k$.}
This hyper-parameter is present in all the methods, except for full attention.
Therefore, we also conduct this ablation for two representative baselines: SWA and Gated DeltaNet.
Not surprisingly, a larger $k$ improves performance for all three methods, as shown in the leftmost panel of Figure~\ref{fig:ablations}, 
and TTT-E2E has similar sensitivity to changes in $k$ compared to the baselines.
We choose $k=8$K as the default since a smaller $k$ does not significantly improve runtime.

\textbf{TTT-E2E with full attention.}
The window size ablation is conducted with only pre-training on DCLM without fine-tuning on Books, so the results above are evaluated on DCLM as well.
Since the pre-training context length is also 8K, SWA with $k=8$K is exactly full attention, and TTT-E2E with $k=8$K becomes the same as TTT-E2E on top of full attention.
It is especially interesting to observe that TTT-E2E can improve the test loss (by 0.018) even on top of full attention, and the difference between TTT-E2E and SWA does not change significantly as $k$ increases.
This observation suggests that TTT-E2E is not merely compensating for the difference between full attention and SWA; 
instead, it produces an orthogonal improvement when other factors, such as context length, are fixed.

\textbf{TTT mini-batch size $b$}.
The middle panel of Figure~\ref{fig:ablations} experiments with the TTT mini-batch size $b$, ranging from 1K to 8K.
This hyper-parameter is unique to methods derived from the TTT perspective, so the only other baseline that allows for a meaningful comparison here is TTT-KVB.\footnote{
Mamba 2 and Gated DeltaNet have a somewhat similar hyper-parameter called chunk size.
However, for these methods, chunk size only affects their hardware utilization rather than their output, so they do not allow for a meaningful comparison.
}
Similar to the window size ablation, the models are evaluated on DCLM after pre-training. 
For both TTT-E2E and TTT-KVB, we observe that a larger choice of $b$ significantly hurts performance. However, a choice of $b$ smaller than 1K also significantly hurts our hardware utilization and stability, to the point that it becomes difficult to experiment with.
Therefore, we choose $b=1$K as the default.

\textbf{Modified architectures without TTT.}
The choice of $b=8$K is equivalent to not doing TTT at all, because our pre-training context length is also 8K.
However, both TTT-E2E and TTT-KVB without TTT are slightly different from Transformer with full attention, because both of these methods have slightly modified the Transformer architecture, as previously illustrated in Figure~\ref{fig:main}.
So do these modifications still matter without TTT?
Figure~\ref{fig:ablations} suggests that the answer is no.
Without TTT, the loss for either TTT-E2E (2.825) or TTT-KVB (2.826) is almost no different from full attention (2.827).
This observation suggests that architecture design plays a minor, supporting role in our method.


\subsubsection{Number of Layers Updated}
We now turn to the most important ablation.
As discussed in Subsection~\ref{subsec:main}, the number of layers updated during TTT controls the amount of storage in which we can compress the information in the context window. 
Therefore, we investigate its effect in terms of context scaling, and present this ablation in the format of Figure~\ref{fig:teaser} (left).
Specifically, for each number of layers, we pre-train a single checkpoint on DCLM and then fine-tune five versions on Books, one for each context length, so the final results are evaluated on Books.\footnote{
In order to fine-tune at 64K and 128K context length, which is quite unusual for a 760M model, we had to double the (outer-loop) batch size given by the basic recipe.
This modification allows us to average over enough sequences so our fine-tuning runs are stable.
For clean comparison, we adopt this modification for fine-tuning at every context length, including the shorter ones, within the scope of this ablation.
}

We experiment with updating the last 1/2, 1/4, and 1/8 of the layers.
For our 760M model with a total of 24 layers, these ratios translate to the last 12, 6, and 3 layers. 
We also experiment with updating only the final layer.
From the rightmost panel of Figure~\ref{fig:ablations}, we observe that when updating only 1 or 3 layers, our method does not scale with context length in the same way as full attention.
When updating 6 or 12 layers, our method does scale.
However, updating 12 layers only performs at roughly the same level as 6.
Therefore, we always update the last 1/4 regardless of model size.


\subsection{Scaling with Training Compute}
\label{subsec:results_train}

\begin{figure}
\vspace{-4ex}
    \centering
    \includegraphics[width=0.495\linewidth]{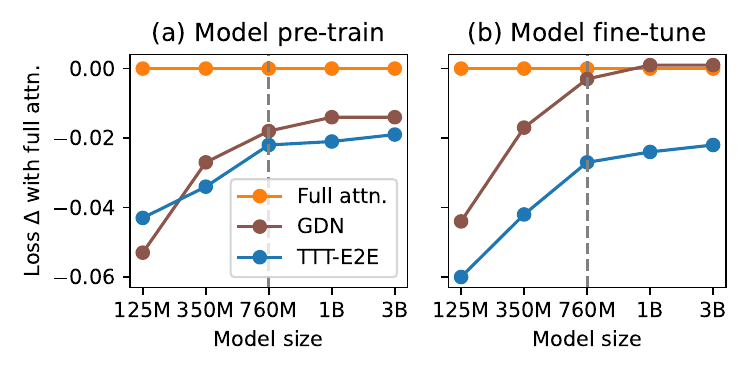}
    \includegraphics[width=0.495\linewidth]{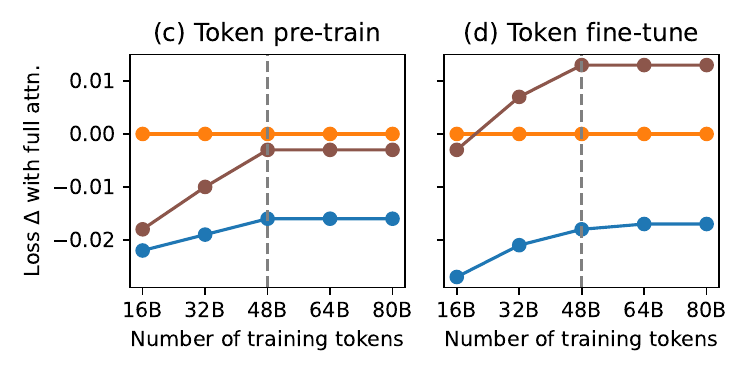}
\vspace{-3ex}
    \caption{
Scaling with training compute in two axes: model size (left) and number of training tokens (right);
    see details in Subsection~\ref{subsec:results_train}.
    Overall, TTT-E2E exhibits a similar trend to full attention under a large training budget (right of the dotted line).
    We report results both on DCLM at 8K context length after pre-training (a, c) and on Books at 32K after fine-tuning with the same context length (b, d).
    Loss $\Delta$ ($\downarrow$), the $y$-value, is the same as in Figure~\ref{fig:teaser} and \ref{fig:ablations}.
    The legend in the leftmost panel is shared across all panels.
    }
    \label{fig:train_scale}
\vspace{-1ex}
\end{figure}

In general, there are two axes of training compute: the model size and the number of training tokens.
We investigate the behavior of our method along these axes when compared to full attention and Gated DeltaNet, and present the results in Figure~\ref{fig:train_scale}.
We choose Gated DeltaNet as the representative among the RNN baselines because it is the most recent work with highly optimized training time.

One popular practice for measuring the effect of training compute is to evaluate on the pre-training dataset immediately after pre-training, as in many scaling law papers~\cite{kaplan2020scaling, hoffmann2022training}.
In the left panels of Figure~\ref{fig:train_scale}, we follow this practice and evaluate on DCLM after pre-training.
But as discussed in Subsection~\ref{subsec:results_ablations}, our window size is the same as the pre-training context length, making SWA, our baseline architecture, equivalent to full attention.
This equivalence raises the concern that the practice discussed above might not reveal the true behavior of our method without full attention.
So we also evaluate on Books at 32K context length after fine-tuning, as shown in the right panels.\footnote{We evaluate at 32K because fine-tuning the smaller models (125M and 350M) is unstable at longer context length.}

For scaling with model size, we simply vary across the five sizes in our basic recipe.
For scaling with the number of training tokens, we keep the model size fixed at 760M, and vary the number of training tokens for pre-training and fine-tuning.
Specifically, our basic number of tokens for pre-training is taken from the Chinchilla recipe~\cite{hoffmann2022training}, and our basic number for fine-tuning is 5\% of that for pre-training, as discussed in Appendix~\ref{app:configs}. 
We experiment with up to $5\times$ the basic number for pre-training and fine-tuning, keeping the 5\% ratio fixed.

\textbf{Similar trend to full attention under large budget.}
We observe a similar trend across the panels:
\vspace{-0.5ex}
\begin{itemize}[itemsep=2pt, topsep=0pt, parsep=0pt, partopsep=0pt]
    \item The advantage of TTT-E2E over full attention visibly decreases with more training compute in the regime of small compute budget. 
    \item However, in the regime of medium compute budget, TTT-E2E follows a similar scaling trend to full attention, as indicated by the blue line staying relatively flat. Although there is still a small uptick for scaling with model size, we expect this uptick to disappear for even larger models given the overall trend.
\end{itemize}
\vspace{-0.5ex}
For scaling with model size, the boundary for the change of regime is roughly 760M.
For scaling with number of training tokens, this boundary is roughly 48B.
We mark these boundaries in Figure~\ref{fig:train_scale} with dotted vertical lines.
It is especially interesting to observe that Gated DeltaNet follows the same trend as TTT-E2E. 
We offer two potential explanations for this observation:
\vspace{-0.5ex}
\begin{itemize}[itemsep=2pt, topsep=0pt, parsep=0pt, partopsep=0pt]
\item Our method can also be interpreted as a hybrid RNN, similar to Gated DeltaNet, as explained in Subsection~\ref{subsec:alternative}. We expect RNNs (sequence models with hidden states of fixed size) to share a similar trend for scaling with training compute.
\item Transformers are widely known to under-perform with insufficient training compute compared to RNNs~\cite{kaplan2020scaling, hoffmann2022training}. Our observations can be interpreted as a deficiency of the full attention baseline with small compute, rather than a deficiency of RNNs with large compute.
\end{itemize}
\vspace{-0.5ex}
Overall, our empirical observations strongly indicate that TTT-E2E should produce the same trend as full attention for scaling with training compute in large-budget production runs.

\textbf{Sensitivity to tokenizer and data quality.}
During our scaling investigation, we collected anecdotal observations on the effect of tokenizer and data quality, as indicated by recency.
Specifically: 
\vspace{-0.5ex}
\begin{itemize}[itemsep=2pt, topsep=0pt, parsep=0pt, partopsep=0pt]
\item Switching to the Llama\,3 tokenizer (2024) from the Llama\,2 tokenizer (2023) improved our advantage over full attention by about 0.01 for 3B models.
\item Switching to DCLM (2024) from SlimPajama (2023)~\cite{cerebras2023slimpajama} enabled our method to produce the same trend as full attention for scaling with number of training tokens after 48B;
our trend with FineWebEdu (2024)~\cite{lozhkov2024fineweb-edu} is also the same as full attention.
With SlimPajama, our lines in the right panels of Figure~\ref{fig:train_scale} exhibited a small uptick, similar to those in the left panels for scaling with model size. 
\end{itemize}
\vspace{-0.5ex}
A comprehensive investigation of these effects would entail reproducing Figure~\ref{fig:train_scale} for a wide variety of tokenizers and datasets, which is beyond the scope of our paper.
Nevertheless, our anecdotal observations might still offer a starting point for future work.
An especially interesting direction is TTT on self-generated tokens, which can be a filtered or rephrased version of the current mini-batch of tokens or a review of the previous mini-batches.
It is widely known that the gating mechanisms in RNNs can guard the hidden states against spurious inputs and better retain the information in valuable ones~\cite{hochreiter1997long, chung2014empirical}. 
We believe that self-generation during TTT can play a similar role. 

\begin{figure}
\vspace{-1ex}
    \centering
    \includegraphics[width=0.95\linewidth]{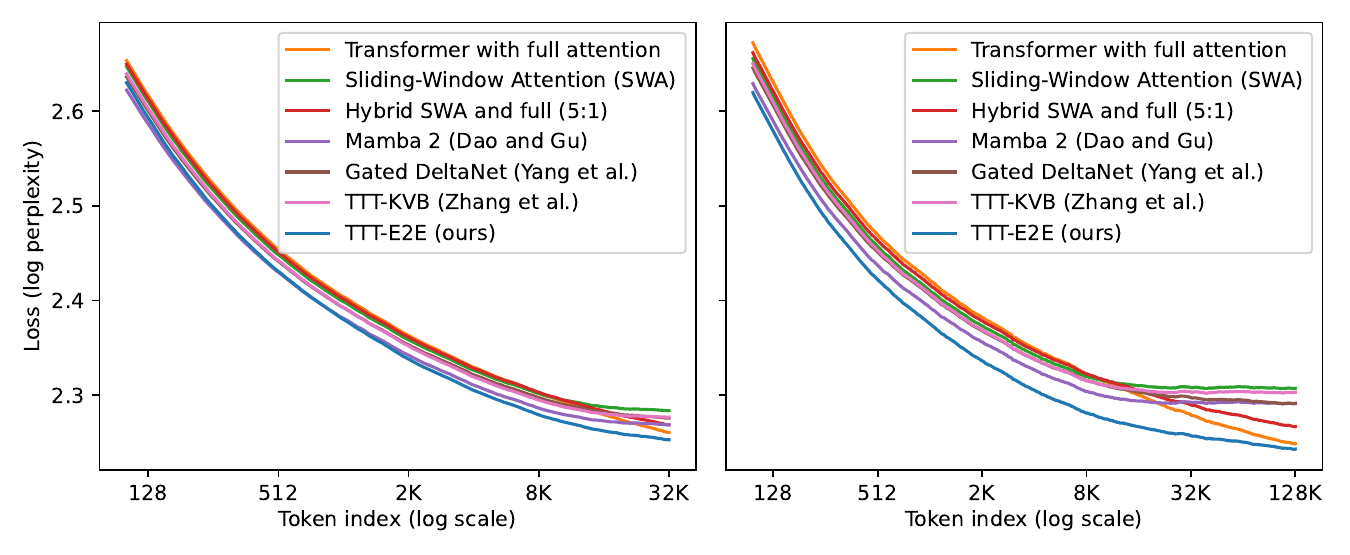}
    \caption{
    Loss breakdown by token index, for context length 32K (left) and 128K (right), following the same process as when we produced the right panel of Figure~\ref{fig:toy}; 
    see details in Subsection~\ref{subsec:results_ctx}.
    Overall, TTT-E2E is the only method that always achieves lower losses than full attention throughout the entire context length,
    and its aggregated advantage mostly comes from the earlier tokens.
    }
    \label{fig:token}
\end{figure}

\subsection{Scaling with Context Length}
\label{subsec:results_ctx}

We presented the key results for scaling with context length in Figure~\ref{fig:teaser} on the first page.
Here, we discuss the setup of these experiments and present a breakdown of some of these results in Figure~\ref{fig:token}.
In addition, Figure~\ref{fig:appendix} in the appendix directly plots the loss values in Figure~\ref{fig:teaser} instead of the loss $\Delta$s.

For the experiments in Figure~\ref{fig:teaser}, we use the largest model (3B) in our basic recipe.
We also use $3\times$ the basic number of tokens for both pre-training and fine-tuning.
As discussed, the basic number for pre-training is taken from the Chinchilla recipe, and that for fine-tuning is 5\% of pre-training.
As in our previous experiments, we pre-train a single checkpoint on DCLM and then fine-tune five versions on Books, one for each context length, so the final results are evaluated on Books.

\subsubsection{Loss Breakdown by Token Index}

Figure~\ref{fig:token} focuses on two context lengths, 32K and 128K, and breaks down the corresponding results in Figure~\ref{fig:teaser} by token index;
we have followed the same process in Subsection~\ref{subsec:toy} to produce the right panel of Figure~\ref{fig:toy}.
Specifically, given a context length $T$, for each $t=1,\dots,T$, we plot the test loss of the next-token prediction task that conditions on $x_0,\dots,x_{t-1}$ and tries to predict $x_t$.\footnote{
We repeat this process for every baseline, but for TTT-E2E in particular, our breakdown lends another interpretation, as discussed in Subsection~\ref{subsec:meta}.
Our test loss at each index $t$ is also the test-time training loss $\ell_t(W_{t-1})$.
Since $W_{t-1}$ has never seen $x_t$, it is fair to compare $\ell_t(W_{t-1})$ with the test loss of approaches without TTT.
}
Therefore, for each method with context length $T$, its test loss in Figure~\ref{fig:teaser}  is the average of all the losses on its corresponding curve in Figure~\ref{fig:token}.
It is important to note that the breakdown for 32K is not a subset of that for 128K, since they are produced from two different models.

We make the following observations from both panels of Figure~\ref{fig:token}:
\vspace{-0.5ex}
\begin{itemize}[itemsep=2pt, topsep=0pt, parsep=0pt, partopsep=0pt]
\item TTT-E2E is the only method that always achieves lower losses than full attention throughout the entire context length.
\item  The difference in test loss between TTT-E2E and full attention is small around the end of the context window.
The aggregated advantage of TTT-E2E over full attention mostly comes from the earlier tokens.
\end{itemize}
\vspace{-0.5ex}
The fact that both observations hold simultaneously for both panels is especially interesting in a somewhat paradoxical way.
As part of the second observation, the difference between TTT-E2E and full attention in the left panel is small around $t=$32K, the end of the context window.
Without other information, one might even speculate that the curves would cross for larger context lengths, such as 128K.
But this speculation is false, as asserted by the first observation from the right panel.
The breakdown plot for 128K better resembles a stretched out version of that for 32K rather than a speculated continuation.
Given that TTT-E2E maintains the same advantage over full attention across context lengths in Figure~\ref{fig:teaser}, this stretching effect should not be surprising.

What gives TTT-E2E an advantage over full attention for the earlier tokens?
Note that this advantage exists even before $t=1$K, when TTT takes the first gradient step on the first (inner-loop) mini-batch.
In other words, before $t=1$K, TTT-E2E and full attention have exactly the same computation graph and only differ in their weights.
So why do the weights of TTT-E2E produce much lower losses?

Here is an intuitive explanation:
The weights of full attention must prepare to be good at all future tokens in the context window.
Such a task can be very hard, because being good at all possible futures limits the model’s capacity to be good at any particular one.
But the weights of TTT-E2E only need to be good at the present mini-batch of tokens, since TTT will produce future weights for the future tokens.
This more focused task should be much easier.
In fact, a key intuition of TTT \mbox{in general}, as we will discuss in Subsection~\ref{subsec:ttt}, is to focus on the present.


\begin{table}[t]
\vspace{-3ex}
\centering
\renewcommand{\arraystretch}{1.2}
\setlength{\tabcolsep}{5pt}
\resizebox{\textwidth}{!}{
\begin{tabular}{l|ccccc|ccccc|ccccc}
\toprule
& \multicolumn{5}{c|}{S-NIAH-1} & \multicolumn{5}{c|}{S-NIAH-2} & \multicolumn{5}{c}{S-NIAH-3} \\
& \multicolumn{5}{c|}{(pass-key retrieval)}
& \multicolumn{5}{c|}{(number in haystack)}
& \multicolumn{5}{c}{(UUID in haystack)} \\
\cmidrule(lr){2-6}\cmidrule(lr){7-11}\cmidrule(lr){12-16}
Method
& 8K & 16K & 32K & 64K & 128K
& 8K & 16K & 32K & 64K & 128K
& 8K & 16K & 32K & 64K & 128K \\
\midrule
Full attention
& \textbf{1.00} & \textbf{1.00} & \textbf{1.00} & \textbf{1.00} & \textbf{0.99}
& 0.99 & \textbf{1.00} & \textbf{1.00} & \textbf{1.00} & \textbf{0.86}
& 0.64 & \textbf{0.64} & \textbf{0.67} & \textbf{0.83} & \textbf{0.64} \\
SWA
& \textbf{1.00} & 0.50 & 0.26 & 0.13 & 0.07
& \textbf{1.00} & 0.43 & 0.28 & 0.16 & 0.05
& 0.57 & 0.41 & 0.24 & 0.09 & 0.05 \\
Hybrid SWA and full
& \textbf{1.00} & 0.93 & 0.88 & 0.69 & 0.21
& \textbf{1.00} & \textbf{1.00} & 0.99 & 0.89 & 0.29
& 0.63 & 0.56 & 0.32 & 0.17 & 0.06 \\
Mamba 2~\cite{dao2024transformers}
& 0.99 & 0.49 & 0.26 & 0.13 & 0.07
& 0.99 & 0.43 & 0.28 & 0.16 & 0.05
& 0.77 & 0.36 & 0.24 & 0.08 & 0.04 \\
Gated DeltaNet~\cite{yang2024gated}
& \textbf{1.00} & 0.50 & 0.26 & 0.13 & 0.07
& \textbf{1.00} & 0.43 & 0.27 & 0.16 & 0.05
& \textbf{0.91} & 0.45 & 0.23 & 0.07 & 0.03 \\
TTT-KVB~\cite{zhang2025test}
& 0.98 & 0.43 & 0.22 & 0.10 & 0.01
& \textbf{1.00} & 0.43 & 0.27 & 0.16 & 0.05
& 0.74 & 0.34 & 0.23 & 0.06 & 0.04 \\
TTT-E2E (ours)
& \textbf{1.00} & 0.46 & 0.24 & 0.13 & 0.06
& 0.99 & 0.43 & 0.28 & 0.16 & 0.05
& 0.77 & 0.44 & 0.24 & 0.10 & 0.03 \\
\bottomrule
\end{tabular}
}
\caption{S-NIAH performance across context lengths, with the best results in bold; see details in Subsection~\ref{subsec:niah}.
Overall, Transformer with full attention dramatically outperforms the other methods, including ours, especially in long context.
This observation, combined with findings from our previous subsections, supports the intuition that the strength of full attention lies in its nearly lossless recall.
}
\label{tab:niah}
\end{table}

\subsection{Needle in a Haystack}
\label{subsec:niah}

The motivation for our method, as discussed in Section~\ref{sec:intro}, was to use longer context to achieve better performance in language modeling without having to recall every detail.
Up to this point, we have focused on evaluations that do not require detailed recall.
Here, we consider a popular evaluation explicitly designed for recall known as Needle in a Haystack (NIAH): The model needs to retrieve a target string (needle) in a passage (haystack), where the target string is distinguished by its clear irrelevance to the rest of the passage.
Specifically, we evaluate all the 3B models fine-tuned at 128K context length, on the three NIAH tasks in RULER~\cite{hsieh2024ruler}.

From Table~\ref{tab:niah}, we observe that Transformer with full attention dramatically outperforms the other methods, including ours, especially in long context.
This observation, combined with findings from our previous subsections, supports the intuition that the strength of full attention lies in its nearly lossless recall.
This strength is inherent to the design of self-attention, which attends to the keys and values of all previous tokens in its cache.
In contrast, the key mechanism in our method is compression, which leaves out seemingly irrelevant details, such as the target string.

\begin{figure}[t]
\centering
\begin{minipage}[t]{0.49\textwidth}
    \centering
    \includegraphics[width=\linewidth]{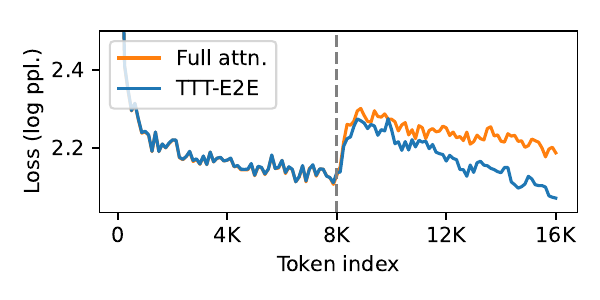}
\end{minipage}\hfill
\begin{minipage}[t]{0.49\textwidth}
    \vspace{-21.5ex}
    \caption{
    Decoding long sequences, using Qwen-8B as the evaluator; 
    see details in Subsection~\ref{subsec:results_decode}.
    For each method, we prefill its context window with 8K tokens from Books, decode another 8K tokens as continuation, and then plot the loss of Qwen-8B by token index, averaged over 512 sequences.
    The dotted line marks the boundary between prefill and decode.
    This plot is in linear scale instead of log scale.
    }
    \label{fig:decode}
\end{minipage}
\end{figure}

\subsection{Decoding Long Sequences}
\label{subsec:results_decode}

Up to this point, all our evaluations have required the model to decode no more than a dozen tokens.
As discussed in the end of Subsection~\ref{subsec:main}, when the decoded tokens have filled a TTT mini-batch, TTT-E2E takes a gradient step on this batch of decoded tokens.
Does this method of ``self-training'' at test time work for decoding long sequences?

In practice, scenarios that require decoding long sequences typically arise either after instruction fine-tuning or during reinforcement learning, e.g., when the model generates long chains of thought.
Therefore, it is inherently challenging to evaluate base models, without the two stages above, in a realistic way.
Since these two stages are beyond the scope of our paper, we make our best effort to evaluate the 3B base models we have trained in Subsection~\ref{subsec:results_ctx}.

For the evaluation in Figure~\ref{fig:decode}, we use Qwen-3-8B-Base~\cite{qwen3technicalreport} as the evaluator. Since our models were trained on Books, we prefill their context windows with 8K tokens from Books, decode another 8K tokens as continuation, and then plot the loss (log likelihood) of Qwen-8B on the concatenated 16K sequence by token index.
While Figure~\ref{fig:token} uses log scale for the $x$-axis, Figure~\ref{fig:decode} here uses linear scale, allowing us to easily compare the trends for prefill and decode.
Additional details of this evaluation are provided in Appendix~\ref{app:decode}.

Similar to our previous observations, TTT-E2E achieves lower Qwen loss than full attention in this limited evaluation.
In addition, we have carefully inspected $\approx 20$ samples of the generated text and found them reasonable.
For both methods, the Qwen loss increases sharply at the boundary between prefill and decode, and then gradually decreases again.
This behavior likely arises because Qwen is initially unfamiliar with the generation style of the evaluated method, but then gradually adapts as more generated content accumulates within its context window.

\subsection{Computational Efficiency}

In Figure~\ref{fig:teaser}, we have presented our inference latency, specifically prefill latency, compared to that of the baselines.
Here, we discuss our setup for measuring prefill latency, and consider two additional axes where computational efficiency is important: decode and training.
In particular, we highlight training latency as a significant limitation of our current implementation and discuss two potential directions for improving it.

\textbf{Setup for prefill latency.}
For each method in the right panel of Figure~\ref{fig:teaser}, we took its corresponding 3B model in the left panel and measured its prefill latency on one H100.
We also took additional steps to optimize the inference latency of the PyTorch baselines, as discussed in Appendix~\ref{app:baselines}.
Following Gated DeltaNet~\cite{yang2024gated}, the latency experiments are performed with a constant number of tokens (128K) per (outer-loop) batch. 
For example, at 128K context length, each batch contains one sequence, and at 8K each batch contains 16 sequences.

\textbf{TTT-E2E only uses standard infrastructure.}
At test time, TTT-E2E can simply use the standard infrastructure optimized for training a regular Transformer.
Specifically, since our hidden state takes the form of regular MLP layers, it can be sharded across GPUs using standard tools with no custom kernel.
In contrast, prior work must fit their hidden states onto the individual chips inside a GPU, which significantly limits their hidden state size.
For example, TTT-KVB~\cite{zhang2025test} must reduce its state size with LoRA, while other prior work, such as Mamba\,2~\cite{dao2024transformers} and Gated DeltaNet~\cite{yang2024gated}, must use a linear hidden state and write custom kernels for efficient memory I/O.

\textbf{Decode latency.}
As discussed in the end of Subsection~\ref{subsec:main}, our method does not perform TTT until the decoded tokens have completely filled a TTT mini-batch. So before reaching a full batch, our decode latency is the same as that of a regular Transformer with SWA.
Once we have a full batch, we need a step of TTT before decoding the next batch of tokens, and our latency for this TTT step is the same as that for prefill.
Altogether, our latency for decoding a long sequence of multiple batches is simply the sum of the two latencies above: that of SWA decode and that of our prefill.
Since both are readily available, we do not report separate measurements for the decode latency of TTT-E2E.

\begin{figure}
\vspace{-3ex}
    \centering
    \includegraphics[width=0.99\linewidth]{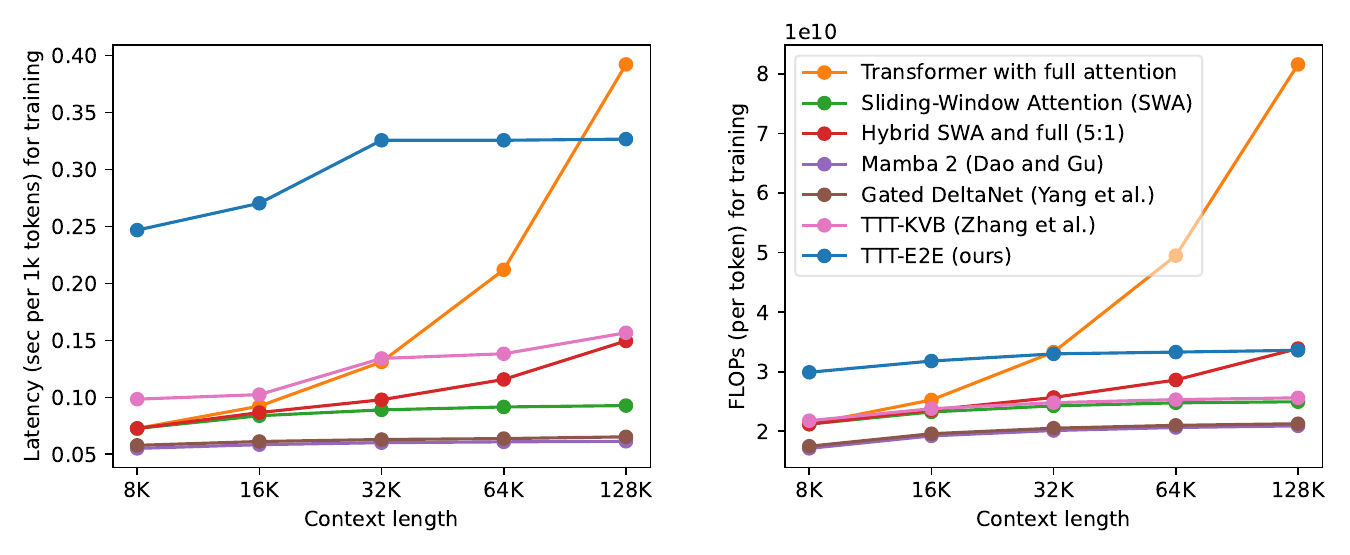}
\vspace{-1ex}
    \caption{
    Training efficiency, in terms of latency on an H200 (left) and FLOPs (right); 
    see details in Subsection~\ref{subsec:results_train}.
    Overall, training latency is still a significant limitation of our current implementation.
    The legend is shared across both panels.
    }
    \label{fig:training_eff}
\vspace{-1ex}
\end{figure}

\textbf{Setup for training latency.}
Most of our training was performed on GB200s.
Since many of our baselines do not have custom kernels written for GB200s (Blackwell), we benchmark training latency on an H200 (Hopper) for fairness to the baselines.
Following our protocol for prefill, we use a constant number of tokens (128K) per batch regardless of context length.

\textbf{Training latency is a limitation.}
At training time, TTT-E2E takes gradients of gradients, which is a much less optimized procedure compared to training a regular Transformer. As shown in the left panel Figure~\ref{fig:training_eff}, our training latency is $1.2\times$ faster than full attention at 128K context length, but $3.4\times$ slower at 8K.
Since most of the training compute is typically spent on pre-training with short context, the training latency of our current implementation remains a significant limitation.
Note that even though our number of FLOPs per token remains constant, as shown in the right panel, our latency grows between 8K and 32K.
This trend arises because we have to increase the amount of gradient checkpointing through time by a factor of $\log(T)$, where $T$ is the context length.\footnote{
By default, libraries such as JAX and PyTorch save the intermediate activations during a forward pass so they can be reused during the backward pass.
However, for a TTT with $W$ as a hidden state, this default saves $W_1,\dots,W_T$, which uses too much memory.
A standard technique to save memory in this scenario is gradient checkpointing~\cite{chen2016training}, which is usually applied through layers, but we apply it through time during training~\cite{sun2024learning, engstrom2025optimizing}.
}

\textbf{Directions for faster training.}
There are two directions for improving our overall training time:
\vspace{-0.5ex}
\begin{itemize}[itemsep=2pt, topsep=0pt, parsep=0pt, partopsep=0pt]
\item Our current implementation cannot use cuDNN FlashAttention~\cite{dao2023flashattention} at training time because it does not support gradients of gradients.
A custom attention kernel would significantly improve our hardware utilization, and potentially eliminate the undesirable trend caused by gradient checkpointing through time.
\item We believe that the training of TTT-E2E can be initialized from a pre-trained Transformer without TTT -- a technique often adopted by prior work on RNNs~\cite{kasai2021t2r, bick2024mohawk, wang2024mambainthellama}.
This practical technique allows TTT-E2E to only take up a small portion of the overall training compute, so the negative effect of its training latency is minimal.
\end{itemize}
\vspace{-0.5ex}
We leave these directions for future work.

\section{Related Work}
\label{sec:related}

\subsection{Continual Learning}

Most of today's AI systems remain static after deployment, even though the world keeps changing. The high-level goal of continual learning is to enable AI systems to keep changing with the world, similar to how humans improve throughout their lives~\cite{hassabis2017neuroscience, de2021continual}.

Conventionally, continual learning as a research field has focused on learning from a \emph{distribution} that gradually changes over time~\cite{lopez2017gradient, van2019three, hadsell2020embracing}.
For example, one could update a chatbot model every hour using new knowledge from the Internet, while typical use cases of the model may require knowledge from both the past and the present~\cite{scialom2022fine, ke2023continual, wang2024comprehensive}.
More formally, at each timestep, we sample new training and test data from the current distribution, update our model using the new training data, and then evaluate it on all the test data up to the current timestep.
Under this setting, most algorithms focus on not forgetting the past when learning from the present~\cite{santoro2016meta, li2017learning, kirkpatrick2017overcoming, gidaris2018dynamic}.



\subsection{Test-Time Training}
\label{subsec:ttt}
The algorithmic framework of test-time training has the same high-level goal as continual learning, but it focuses on two aspects where human learning stands out from the forms of continual learning in the conventional literature.

First, each person has a unique brain that learns within the context of their individual life. This personalized form of continual learning is quite different from, for example, the chatbot model that is fine-tuned hourly using the latest information available worldwide. While such a model does change over time, it is still the same at any given moment for every user and every problem instance.

Second, most human learning happens without a boundary between training and testing. Consider your commute to work this morning. It is both "testing" because you did care about getting to work this very morning, and "training" because you were also gaining experience for future commutes. But in machine learning, the train-test split has always been a fundamental concept.

The concept of test-time training is introduced to realize these two special aspects of human learning.
\emph{Training} typically involves formulating a learning problem (such as empirical risk minimization) and then solving it.
Following \cite{sun2023learning}, \emph{test-time training} is defined as any kind of training that formulates a potentially different learning problem based on each individual test instance.

This concept has a rich history in AI.
A well-known example in NLP is dynamic evaluation, pioneered by Mikolov et al.\,\cite{mikolov2013efficient} and extended by Krause et al.\,\cite{krause2018dynamic}, which our Subsection~\ref{subsec:toy} builds upon.
In computer vision, early examples have also emerged in applications such as face detection~\cite{jain2011online}, video segmentation~\cite{mullapudi2018online}, super-resolution~\cite{shocher2018zero}, and 3D reconstruction~\cite{luo2020consistent}.
Next, we discuss three popular forms of test-time training today, with an emphasis on their connections to each other and to historical examples.

\vspace{-1ex}
\subsubsection{TTT on Nearest Neighbors: Larger Effective Capacity}

One simple form of test-time training was called locally weighted regression in the 1970s~\cite{stone1977consistent, cleveland1979robust}, local learning in the 1990s~\cite{bottou1992local}, and KNN-SVM in the 2000s~\cite{zhang2006svm}: Given a test instance, find its nearest neighbors in the training set, and then train (or fine-tune) the model on these neighbors before making a prediction.
This procedure can significantly increase the effective capacity of the model; for example, it allows a linear model to fit a highly nonlinear ground truth~\cite{stone1977consistent}.

This simple form captures one of the key intuitions of test-time training.
In the conventional view of machine learning, a model, once trained, no longer changes at test time. 
As a consequence, it must prepare to be good at all possible inputs in the future. 
This task can be very hard, because being good at all possible futures limits the model’s capacity to be good at any particular one.
But only one future is actually going to happen.
So why not train our model once this future happens?

Recently, \cite{hardt2023test} extended this idea to modern language models and observed a similar benefit of larger effective model capacity after test-time training, and \cite{hubotter2024efficiently} further improved these results through better strategies for neighbor selection. In addition, \cite{hubotter2025learning} showed that test-time training on neighbors from the training set is also effective with RL for reasoning tasks, and \cite{bagatella2025test} developed the same idea for visual-motor tasks.

\vspace{-1ex}
\subsubsection{TTT for Novel Instances: Better Generalization}
As models become larger today, their competence is often limited not by their capacity, but by the amount of available training data, especially when they need to generalize to novel test instances that are ``out-of-distribution''.
In this case, it is even harder to prepare for all possible test instances in the future, especially the novel ones, with a static model.
But once a specific test instance is given, we can use it to generate relevant data, which we can then use for training~\cite{sun2020test}.
In other words, the ``neighbors'' for TTT do not have to come from the training set; they can also be generated on-the-fly.

Since the test instance is unlabeled, one way to make it useful for training is through self-supervision, which generates new pairs of inputs and labels for an auxiliary task such as masked reconstruction (e.g., BERT~\cite{devlin2018bert} and MAE\cite{mae}).
While the auxiliary task is different from the main prediction task, improving performance in one can help the other through their shared representations.
This form of TTT can significantly improve generalization under distribution shifts~\cite{sun2020test, ttt-mae}.

Recently, TTT has been an important part of AlphaProof~\cite{alphaproof}, which achieved IMO silver-medal standard in 2024.
Given each test problem, their system first generates a targeted curriculum of easier problems by prompting a language model, and then performs reinforcement learning on the generated data.
Another recent work, Akyurek et al.\,\cite{akyurek2024surprising}, found TTT effective for few-shot reasoning tasks such as ARC-AGI. Their system generates augmentations of the few-shot demonstrations in the test problem then performs supervised learning.

\vspace{-1ex}
\subsubsection{TTT on Sequences: Longer Memory}

In all the forms of TTT discussed so far, the model is reset after each prediction because the test instances are independent.
However, humans do not constantly reset their minds.
Our memory of how to solve the previous learning problem often helps with the current one, because our experience in the world is much closer to a correlated sequence of data than independent ones.

Sequential applications, such as videos and robotics, offer a playground that bridges this difference.
For example, \cite{hansen2020self} extended TTT with self-supervision to a manipulation policy whose input is a video stream of the robot's workstation, and found that no reset leads to a much larger improvement.
Recently, \cite{wang2023test} extended the same idea to video segmentation using a model trained with only images.
In this case, TTT can be viewed as compressing the context from previous frames into the weights of the model without learning to learn, similar to the naive version of our method in Subsection~\ref{subsec:toy}.

\textbf{TTT-KVB.}
Text, like videos, is a form of sequence.
In Subsection~\ref{subsec:alternative}, we have discussed TTT-KVB as the most relevant line of prior work~\cite{sun2024learning, zhang2025test, dalal2025one}, which includes variants such as MesaNet~\cite{von2025mesanet}, Titans~\cite{behrouz2024titans}, and Nested Learning~\cite{behrouz2025nested}.
The popularity of TTT-KVB has two side effects:
\vspace{-0.5ex}
\begin{itemize}[itemsep=2pt, topsep=0pt, parsep=0pt, partopsep=0pt]
\item Because the KVB objective is inspired by self-attention, which stores the keys and values, many think that long-context TTT is about memorization instead of generalization.
\item Because TTT(-KVB) layers are drop-in replacements for self-attention layers, many also think of long-context TTT as an approach to architecture design.
\end{itemize}
Our work shows that long-context TTT does not need to memorize the association between the keys and values.
In addition, our method is derived purely under the formulation of a continual learning problem, with minimal changes to the architecture.


\subsection{Fast Weights and Fast Weight Programmers}

The general idea of \emph{fast weights} is to update the parameters of a ``fast'' model on only the most relevant data, as opposed to the conventional practice of updating a ``slow'' model on all data~\cite{tieleman2009using}.
This idea has existed since the 1980s~\cite{feldman1982dynamic, hinton1987using, von1994correlation}.
Because the most relevant data can often include the test instance itself, test-time training can be viewed as a special case of fast weights, with a heavier emphasis on the formulation of an explicit learning problem.

The general idea of \emph{fast weight programmers} (FWPs) is to update the fast weights at test time with a ``slow'' model (as a programmer) that, in turn, is updated less frequently, if at all~\cite{schmidhuber1992learning}.
In our method, the inner-loop weights $W$ can be viewed as ``fast'' and the outer-loop weights $\theta$ as ``slow''.
Therefore, our method can be viewed as a special case of FWPs~\cite{kirsch2021meta}.
Next, we briefly review some of the literature on FWPs in the order of relevance.

\textbf{Clark et al.~\cite{clark2022meta}.} 
This work is the most relevant to ours in methodology.
Given a Transformer baseline with full attention, they add an MLP layer as fast weights, whose initialization is trained as slow weights along with the rest of the model.
Similar to ours, their method updates the fast weights by taking a gradient step on the next-token prediction loss computed over each chunk (mini-batch) of tokens.
Their method significantly improves perplexity compared to the baseline but does not improve efficiency, since their combined architecture does not have linear complexity.
In addition, their design adds the fast weights only to the end of the model instead of interleaving them with attention layers.
In our experiments, interleaving proves to be critical for maintaining the performance gain on top of larger baselines.
Nevertheless, we find Clark et al. to be a valuable inspiration.
An earlier work~\cite{wolf2018meta} also contains sketches of a similar idea with limited experiments.

\textbf{FWPs for long context.}
Many methods addressing the problem of long context have roots in the literature of FWPs.
In particular, \cite{schmidhuber1992learning} (Schmidhuber, 1992) has been a major source of inspiration for
modern RNN layers, such as linear attention~\cite{katharopoulos2020transformers, schlag2020learning}, DeltaNet~\cite{schlag2021linear, yang2024parallelizing}, and
Gated DeltaNet~\cite{yang2023gated}, one of our baselines.
In addition, some of the work on TTT for long context \cite{sun2024learning, zhang2025test} (discussed in Subsection~\ref{subsec:ttt}) can also be viewed as FWPs, due to the connection between TTT and fast weights.
Notably, one instantiation in Irie et al.~\cite{irie2021going} uses MLPs as layer-wise fast weights for long context, preceding the similar instantiation in \cite{sun2024learning}. 

\textbf{Other FWPs.}
While the FWPs above can be interpreted through TTT, many other varieties cannot.
For example, \cite{irie2022modern} designs the fast weights to be programmed by themselves, 
\cite{irie2022images} builds an image generator using the images as fast weights,
\cite{irie2022neural} applies continuous-time extensions of FWPs to time-series classification,
while 
\cite{irie2023practical} and \cite{grazzi2024unlocking} demonstrate how the choice of update rules affects the expressiveness of FWPs on formal language recognition tasks.
In fact, all networks with some gating mechanism, such as Transformers with SwiGLU blocks~\cite{shazeer2020glu}, can also be viewed as FWPs~\cite{gu2023mamba}.

\subsection{Learning to Learn}

For decades, researchers have been arguing that learning to learn, also known as meta-learning or bi-level optimization, should be an important component of intelligence~\cite{schmidhuber1987evolutionary, bengio1990learning, thrun1998learning, lake2017building}.
Perhaps the most relevant work in this field is MAML~\cite{finn2017model}. 
Similar to our work, MAML also has an outer loop that learns the initialization of the inner loop through gradients of gradients.
The main difference between MAML and our work lies in the problem setting.
Specifically, their inner loop learns from an entire dataset at a time, so the outer loop requires a large collection of datasets.
In contrast, our work addresses the problem of language modeling by casting it as learning to learn.
In principle, any supervised learning problem can be cast into our problem formulation.

\section{Conclusion}
\label{sec:discussion}

We have introduced TTT-E2E, a general method for long-context language modeling.
In principle, TTT can be applied to any baseline architecture.
For our experiments, this baseline is a Transformer with sliding-window attention.
Adding our method to this baseline induces a hierarchy often found in biological memory, where the weights updated at test time can be interpreted as long-term memory and the sliding window as short-term memory.
We believe that these two classes of memory will continue to complement each other, and stronger forms of short-term memory will further improve the combined method.



\clearpage

\section*{Author Contributions}

We state the contributions of each of the six core contributors.

\textbf{Arnuv Tandon} and \textbf{Karan Dalal} led the investigations into scaling with training compute and scaling with context length, developed the codebase, conducted the final experiments, managed our cluster, and played a central role in every aspect of this research, including its overall direction.

\textbf{Xinhao Li} developed and conducted most of the early experiments, including the toy experiments, 
co-developed our early codebase with Marcel,
and contributed features for large-scale training.

\textbf{Daniel Koceja} led a set of mid-project experiments that brought clarity to the team's investigation into scaling with training compute.

\textbf{Marcel R{\o}d} developed most of the early codebase and provided expertise to improve latency.

\textbf{Yu Sun} served as the project lead, making decisions on most day-to-day matters and on the project's overall direction. He developed the idea of TTT-E2E, designed most of the early experiments, and established the experimental protocols for the team.
He also wrote the paper.

Yu Sun started the project in October 2024 together with Xinhao Li, Karan Dalal, and Daniel Koceja. Arnuv Tandon and Marcel R{\o}d joined the project in November 2024. Karan Dalal and Daniel Koceja left from November 2024 through March 2025, and rejoined the project in April 2025. Marcel R{\o}d left the project in May 2025. Xinhao Li and Daniel Koceja left in September 2025.

\section*{Acknowledgements}

XW was supported, in part, by NSF CAREER Award IIS-2240014.
TH was supported by the Tianqiao and Chrissy Chen Foundation and a grant under the NSF CAREER IIS-2338866, ONR N00014-24-1-2609, and DARPA Cooperative Agreement HR00112520013. This work does not necessarily reflect the position or policy of the government, and no official endorsement should be inferred.

The authors would like to thank Hyperbolic Labs and SF Compute for their GPU cloud services, 
Matt Behrens and the service team at Voltage Park for their help with our cluster, 
Zacharie Bugaud and Alan Zheng at Astera for discussions during the weekly meetings,
Jan Kautz at NVIDIA for general research support,
Zhuang Liu for discussions during the early phase of the project,
and Arjun Vikram, 
Gashon Hussein, and Aaditya Prasad for their short-term contributions.
Arnuv and Karan would like to thank Harj Taggar at Y-Combinator for his support with compute credits.
Yu would like to thank Songlin Yang and Tianyuan Zhang for their feedback on drafts and results, Mert Yuksekgonul and Chloe Hsu for their support at every phase of the project, as well as his PhD advisors, Alexei A. Efros and Moritz Hardt, for their
many insights from years ago that eventually became part of this paper.

\clearpage

\begin{small}
\bibliographystyle{plain}
\bibliography{refs}
\end{small}

\clearpage
\appendix

\section{Recipe for the Toy Example}
\label{app:toy}

There are two architectures in our toy example: Transformer with full attention, and Transformer without attention.
Their only difference is that the latter has every attention layer removed, therefore fewer parameters.
All the TTT methods are based on the latter architecture.
Both architectures consist of two Transformer blocks.
These blocks are constructed based on those in the 125M model in the basic recipe, with half the embedding dimension (384) and number of attention heads (6).

We train all the methods on DCLM at context length 128.
The number of training tokens follows the Chinchilla recipe, and the (outer-loop) batch size is 125K tokens.
We evaluate on the same held-out partition of DCLM as in Section~\ref{sec:results}.
For each method, we sweep for its optimal learning rate in the range of $[2\mathrm{e}{-3},\,6\mathrm{e}{-3}]$ in increments of $0.5\mathrm{e}{-3}$. 
The optimal learning rate is $3\mathrm{e}{-3}$ for Transformer with full attention, and
$5\mathrm{e}{-3}$ for Transformer without attention and all the TTT variants.

\section{Basic Recipe}
\label{app:configs}

\renewcommand{\arraystretch}{1.45}
\begin{table*}
\vspace{-4ex}
\centering
\small
\begin{tabular}{lcccccccccc}
\toprule
 & \multicolumn{3}{c}{Model configurations}
 & \multicolumn{3}{c}{Pre-training recipe}
 & \multicolumn{3}{c}{Fine-tuning recipe} \\
\cmidrule(lr){2-4} \cmidrule(lr){5-7} \cmidrule(lr){8-10}
Params. & Blocks & Dim. & Heads
& Batch size & LR & Total size
& Batch size & LR & Total size\\
\midrule
125M & 12 & 768  & 12
& 0.5M  & 3e-3   & 2.5B
& 1M   & 4e-4   & 125M \\
350M & 24 & 1024 & 16
& 0.5M & 1.5e-3 & 7B
& 1M   & 4e-4   & 350M \\
760M & 24 & 1536 & 16
& 0.5M & 1.25e-3 & 15B
& 1M   & 4e-4   & 750M \\
1.3B & 24 & 2048 & 32
& 0.5M & 1e-3   & 26B
& 1M  & 4e-4   & 1.3B \\
2.7B & 32 & 2560 & 32
& 1M & 8e-4   & 54B
& 2M  & 4e-4   & 2.7B \\
\bottomrule
\end{tabular}
\caption{
Our basic recipe, where
dim.~means the embedding dimension, batch size means the number of tokens per (outer-loop) batch, and
total size means the total number of tokens in the training set.
Since training only has one epoch, the number of training steps is total size / batch size.
}
\label{tab:configs_combined}
\end{table*}

We use the standard Transformer architecture with QK norm~\cite{olmo2025olmo3}, and the Llama\,3 tokenizer~\cite{dubey2024llama}.
Our basic recipe is shown in Table~\ref{tab:configs_combined}.
It uses the model configurations and pre-training recipe in GPT-3~\cite{brown2020language}, with two changes following the Mamba\,2 paper~\cite{dao2024transformers}:
We use $5\times$ the peak learning rate in GPT-3, and half the batch size for the 1B model.
These two changes have been shown to improve the full attention baseline.
Our learning rate schedule, also following Mamba\,2, is the same for every model size:
For the first 10\% of training, our learning rate increases from 0 to the peak in a linear schedule. 
For the rest of training, it decays to 1e-5 in a cosine schedule.

For extension fine-tuning, we always use 5\% of the number of pre-training tokens.
We also double the batch size so there can be a reasonable number of sequences per batch.
To reduce the cost of hyper-parameter search, we only sweep the peak learning rate for the full attention baseline.
Specifically, for each model size in [760M, 1B, 3B], and each context length in [32K, 128K], we sweep the following learning rates: 
[$8\mathrm{e}{-5},\, 1\mathrm{e}{-4},\, 2\mathrm{e}{-4},\, 4\mathrm{e}{-4},\, 8\mathrm{e}{-4}$].
We find that a single choice, $4\mathrm{e}{-4}$, happens to perform the best across all the model sizes and context lengths above.
Therefore, we use this learning rate for every context length and model size in the basic recipe.
Following~\cite{liu2024world}, we restart the cosine schedule at the beginning of fine-tuning.

We use RoPE~\cite{kitaev2020reformer} with $\theta=500$K for pre-training at 8K context length, following Llama~3~\cite{dubey2024llama}. 
For extension fine-tuning, we change the RoPE $\theta$ for full attention, following standard practice~\cite{dubey2024llama, xiong2023effective}. 
Unfortunately, we could not find the values of $\theta$ in most of the papers on long context.
Following~\cite{liu2025worldmodelmillionlengthvideo}, we use $\theta=10$M at 128K, and choose the other values of $\theta$ by assuming a log-linear relationship with context length: $\theta=1$M for 16K, $2$M for 32K, and $5$M for 64K.

\section{Improvements to the Baselines}
\label{app:baselines}

We made two improvements to the baselines.

\textbf{Latency.}
Our method, implemented in JAX, uses the latest cuDNN kernel for attention.
However, all the RNN baselines, implemented in PyTorch, originally used the FlashAttention\,2 kernel~\cite{dao2023flashattention}, which was much less efficient.
To improve the baselines, we upgraded their attention layers to use the latest version of FlashAttention~3~\cite{shah2024flashattention3fastaccurateattention}.

\textbf{QK norm.}
We find that normalizing the queries and keys in the attention layers (QK norm) makes training more stable for TTT-E2E, and also improves the Transformer baselines, as suggested by prior work~\cite{olmo2025olmo3}.
Therefore, we apply it to the other baselines with sliding-window attention layers. 
At the scale of 760M, applying QK norm to Gated DeltaNet improves its loss from 2.814 to 2.809 for pre-training at 8K context length, and from 2.691 to 2.683 for extension fine-tuning at 32K.

\section{Additional Details for Decoding Evaluation}
\label{app:decode}

For sampling, we set the softmax temperature to 1, and the threshold $p$ for top-$p$ vocabulary to 0.95, following~\cite{holtzman2019curious}.

In addition, it is widely known that base models without instruction fine-tuning tend to decode patterns of repetition.
To address this problem, we use the standard API for repetition penalty in the HuggingFace toolkit for generation, and set its value to 1.1. 
This value is recommended by many HuggingFace users, and makes the full attention baseline generate reasonable text, as we have found through manual inspection.

\begin{figure}[t!]
\vspace{-2ex}
    \centering
    \includegraphics[width=0.7\linewidth]{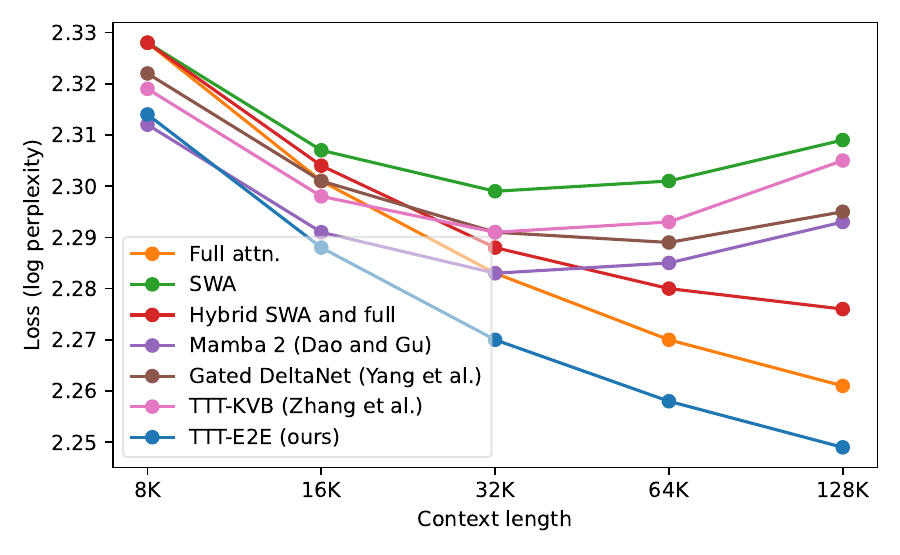}
    \caption{Scaling with context length. We use the same results as in the left panel of Figure~\ref{fig:teaser}, but directly plot the loss values instead of the loss $\Delta$s.
    Longer context length improves loss for full attention and hybrid SWA and full across all the context lengths.
    It also improves loss for every method up to 32K.
    However, for SWA, Mamba~2, Gated DeltaNet and TTT-KVB, longer context length hurts loss after 32K. 
    This trend arises because for longer context, there are fewer training sequences per (outer-loop) mini-batch during extension fine-tuning, so the gradients have higher variance; at the same time, these methods cannot effectively leverage the benefit of longer context, so the harm of the higher variance outweighs the benefit.
    }
    \label{fig:appendix}
\end{figure}

\end{document}